\documentclass{article}
\usepackage{ijcai19}

\usepackage{times}
\usepackage{soul}
\usepackage[utf8]{inputenc}
\usepackage[small]{caption}
\usepackage{amsfonts}
\usepackage{graphicx}
\usepackage{bmpsize}
\usepackage{amsmath}
\usepackage{booktabs}
\usepackage{algorithm}
\usepackage{algorithmic}
\usepackage{color}
\usepackage{multirow}
\usepackage{subfigure}
\usepackage{breqn}

\newcommand{\defmodel}{\textbf{DefNet}~}
\newcommand{\defmodele}{\textbf{DefNet}}
\newcommand{\gnn}{GNN~}
\newcommand{\acl}{\textbf{ACL}~}
\newcommand{\ger}{\textbf{GER}~}

\newcommand{\shen}[1]{\textcolor{red}{(shen: #1)}}
\newcommand{\zach}[1]{\textcolor{green}{(zach: #1)}}
\newcommand{\jni}[1]{\textcolor{blue}{(jni: #1)}}
\newcommand{\nop}[1]{}

\title{Adversarial Defense Framework for Graph Neural Network}

\author{Shen Wang\thanks{University of Illinois at Chicago, \{swang224, psyu\}@uic.edu.}
\and
Zhengzhang Chen\thanks{NEC Laboratories America, \{zchen, jni, xiao, zhichun, haifeng\}@nec-labs.com.} \thanks{Corresponding author.}
\and
Jingchao Ni\footnotemark[2]
\and
Xiao Yu\footnotemark[2]
\and\\
Zhichun Li\footnotemark[2]
\and
Haifeng Chen\footnotemark[2]
\and
Philip S. Yu\footnotemark[1] 
}
\begin{document}

\maketitle

\begin{abstract}
Graph neural network (GNN), as a powerful representation learning model on graph data, attracts much attention across various disciplines. However, recent studies show that GNN is vulnerable to adversarial attacks. How to make GNN more robust? What are the key vulnerabilities in GNN? How to address the vulnerabilities and defense GNN against the adversarial attacks? In this paper, we propose \defmodele, an effective adversarial defense framework for GNNs. In particular, we first investigate the latent vulnerabilities in every layer of GNNs and propose corresponding strategies including dual-stage aggregation and bottleneck perceptron. Then, to cope with the scarcity of training data, we propose an adversarial contrastive learning method to train the \gnn in a \textit{conditional GAN} manner by leveraging the high-level graph representation. Extensive experiments on three public datasets demonstrate the effectiveness of \defmodel in improving the robustness of popular \gnn variants, such as \textit{Graph Convolutional Network} and \textit{GraphSAGE}, under various types of adversarial attacks. 
\end{abstract}
\vspace{-10pt}
\section{Introduction}
\label{sec:intro}
Graph Neural Network (GNN) has received wide attention~\cite{defferrard2016convolutional,hamilton2017inductive,velickovic2017graph,kipf2016semi,li2015gated,gilmer2017neural} in the past few years. It extends the traditional neural networks for processing the data represented in the graph domains including social networks, protein-protein interaction network, information network, and \textit{et al}. The goal of \gnn is to learn the representation of the graph, in the node level or graph level, via a neural network consisting of a neural encoding function and a decoding function. Because of its remarkable graph representation learning ability, \gnn has been explored in various real-world applications, such as physics system \cite{Battaglia2016}, financial system \cite{Liu2018}, protein interface prediction \cite{Fout2017ProteinIP}, and disease classification \cite{ijcai2018490}.

However, recent studies~\cite{dai2018adversarial,zugner2018adversarial} have found that GNNs can easily be compromised by adversarial attacks. These adversarial attacks are often stealthy and only require small perturbations (\textit{e.g.}, by adding or dropping edges) of the graph structure and/or node features to induce the GNNs to make incorrect predictions for some specific nodes with high confidence. This makes it highly risky for those aforementioned domains that \gnn applied, especially security or financial applications, because it could potentially open the backdoor for the attackers.\nop{, because small perturbations are easy to inject by the adversaries. For example, faudsters could try to disguise themselves by connecting other normal users.}

\nop{Despite the recent advances in adversarial attacks on GNN, currently there is no study on how to defend the \gnn against the adversarial attacks.}

Despite recent advances in adversarial attacks~\cite{dai2018adversarial,zugner2018adversarial,goodfellow2015explaining},\nop{and DNN~\cite{szegedy2013intriguing,goodfellow2015explaining},} the question of how to defend the \gnn against the adversarial attacks has not been satisfyingly addressed yet.\nop{currently there is no study on how to defend the \gnn against the adversarial attacks.}\nop{the question on how to defend the adversarial attacks has not been satisfyingly addressed yet.} Nonetheless, some works have been proposed to improve the robustness of traditional Deep Neural Networks (DNNs)~\cite{goodfellow2015explaining,madry2017towards,na2017cascade,samangouei2018defense,papernot2016distillation,jia2017adversarial}. Among them, adversarial learning, which augments the training data with adversarial examples during the training stage~\cite{goodfellow2015explaining,madry2017towards,na2017cascade}, has shown to be most effective in defending the powerful adversarial attacks~\cite{athalye2018obfuscated}.\nop{They fall into three main categories: (1) adversarial training methods \cite{goodfellow2015explaining,madry2017towards,na2017cascade},\nop{which augments the training data with adversarial samples to improve the robustness of the classifier}; (2) adversarial perturbation removing methods \cite{samangouei2018defense}; and (3) smoothness enforced methods \cite{papernot2016distillation}.} However, these defense works only focus on either image data~\cite{samangouei2018defense} or text data~\cite{jia2017adversarial}.\nop{in which the data follow an i.i.d distribution and lie in the continuous feature space.} When it comes to \textit{adversarial defense in graph data}, where the graph structure and the nodes' feature are often discrete, there is no existing work. It is difficult to defend the \gnn against the adversarial attacks due to two major challenges: (1) the latent vulnerabilities from every layer of \gnn and (2) the discrete adversarial space.

Naturally, the vulnerabilities of \gnn can come from every layer of its unique architecture: the aggregation layer and the perceptron layer. As the main strength of GNN, the aggregation layer computes the node representation by leveraging its context, which covers the graph structure. However, the aggregation layer can also be vulnerable\nop{is also an influential vulnerability. This is} due to the effect that, the node representation depends not only on its own feature but also on its neighborhood \cite{kipf2016semi}. \cite{zugner2018adversarial} has shown that the attack can be conducted without touching of the target node, since the attack happening on the other neighbors may propagate to the nodes that are not attacked. 
Another vulnerability is related to the input dimension, which inherits from DNNs. As shown in \cite{simon2018adversarial}, there is a one-to-one relationship between the adversarial vulnerability and the input dimension, such that the adversarial vulnerability increases with the input dimension. \nop{By noticing the vulnerabilities caused by different components, it requires to improve their robustness to adversarial attack.}

In addition to the lack of understanding of the vulnerabilities, the discrete adversarial space is also an issue.\nop{Existing works on adversarial defense only concentrate the on the image, which consists of continuous features, while} Different from images where the data are continuous, the graphs are discrete and the combinatorial nature of the graph structures makes it much more difficult than the text data. Thus, how to generate good adversarial graph samples to augment the training data for \gnn is a non-trivial problem. Simply generating adversarial samples from a noise distribution will not tailor toward the graph data. And the generated ineffectual adversarial samples could even weaken the robustness of the model against various adversarial attacks.

To tackle the aforementioned two challenges, in this paper, we propose \defmodele, an effective framework for \underline{def}ending popular Graph Neural \underline{Net}works against adversarial attacks. \defmodel consists of two modules: \textbf{GER} (Graph Encoder Refining) and \textbf{ACL} (Adversarial Contrastive Learning).
First, \ger investigates the vulnerabilities in the aggregation layer and the perceptron layer of a \gnn encoder, and applies dual-stage aggregation and bottleneck perceptron to address those vulnerabilities. Then, in \acl module, the \gnn is trained in an adversarial contrastive learning style. To overcome the discrete adversarial space problem, \acl models the adversarial learning as a \textit{conditional GAN} by leveraging the high-level graph representation as auxiliary information to regularize the node representation learning.\nop{As a result, the trained \gnn become more robust to the adversarial samples.} To evaluate the performance of \defmodele, we perform an extensive set of experiments on three public graph datasets. The results demonstrate
the effectiveness of our proposed framework on defending the popular \gnn variants, such as Graph Convolutional Network and GraphSAGE, against various types of adversarial attacks.

\vspace{-3pt}
\section{Preliminaries and Problem Formulation}
\label{sec:Pri}

\nop{
}

In this paper, we use bold lowercase for vectors (\emph{e.g.}, $\mathbf{a}$), bold capital for matrices (\emph{e.g.}, $\mathbf{A}$), and calligraphic letters for sets (\emph{e.g.}, $\mathcal{A}$). A graph is represented by a triplet $G=(\mathcal{V}, \mathcal{E}, \mathbf{X})$, where $\mathcal{V} = \{v_1, ..., v_n\}$ is the set of nodes, $\mathcal{E} \subseteq \mathcal{V} \times \mathcal{V}$ is the set of edges, and $\mathbf{X} \in \mathbb{R}^{n \times d}$ is a matrix with the $v$-th row, $\mathbf{x}_{v}$, representing the $d$-dimensional feature vector of node $v$. Also, $|\mathcal{V}| = n$ and $|\mathcal{E}| = m$ are the number of nodes and edges in $G$, respectively. Following the existing work on adversarial attacks~\cite{zugner2018adversarial}, in this paper, we consider graphs that are undirected and attributed\nop{and the Graph Neural Network (GNN) is convolutional aggregator based~\footnote{Attention based GNNs like Graph Attention Network or Gate updater based GNNs like Graph LSTM won't covered by this work.}}. 





\nop{Next, we given the definition of a graph neural network and its important components. }

\nop{\subsection{Preliminaries}}
\medskip
A \textbf{Graph Neural Network} (GNN) is a function $f_{GNN}$, parameterized by neural networks. Typically, $f_{GNN}$ is composed of an encoding function $ENC(\cdot)$ and a decoding function $DEC(\cdot)$, such that $f_{GNN}(G) \equiv DEC(ENC(G))$. In particular, we have
\begin{align}
\mathbf{h}_v &= ENC (\mathbf{x}_v, \{\mathbf{x}_u\}_{u \in \mathcal{N}_{v}})\\
\mathbf{o}_v &= DEC (\mathbf{h}_v)
\end{align}
where $\mathbf{h}_v$ is the embedding vector of node $v$, $\mathbf{o}_v$ is the decoded output, and $\mathcal{N}_{v}$ is the set of neighboring nodes of $v$ in graph $G$, \textit{i.e.}, $\mathcal{N}_{v} = \{u | (u, v) \in \mathcal{E}, (v, u) \in \mathcal{E}\}$. \nop{\jni{directed or undirected}}

\nop{Moreover,}The encoding function is characterized by 
an {\em aggregation layer} and a {\em perceptron layer} \cite{hamilton2017representation}. The aggregation layer aims to aggregate the neighborhood information of a node for updating its embedding, which encourages message propagation along the graph. Formally, the aggregation function is defined as follows:
\begin{align}
\mathbf{a}_{v}^{(k)} &= AGG( \{ \mathbf{h}_{u}^{(k-1)}\}_{u \in \mathcal{N}_v} )
\end{align}
where $\mathbf{a}_{v}^{(k)}$ is the aggregated representation of the neighboring embeddings $\mathbf{h}_{u}^{(k-1)}$ (at the ($k-1$)-th layer) of node $v$.

After obtaining $\mathbf{a}_{v}^{(k)}$, the perceptron layer transforms it to a hidden nonlinear space to reach an updated embedding, $\mathbf{h}_v^{(k)}$, of node $v$ by performing:
\begin{align}
\mathbf{h}_v^{(k)} = PERCE (\mathbf{a}_v^{(k)})
\end{align}

\nop{The perceptron layer consists of a linear mapping and a nonlinear gating function, with shared parameters across all the nodes. Benefit from its weight sharing, the model has less parameters, and improved efficiency. Meanwhile, the weight sharing facilitates a regularization to preserve the invariant property in the graph.\jni{can we remove this paragraph?}}

Depending on different tasks, the parameters of $f_{GNN}$ can be trained in a supervised or unsupervised manner. For example, considering node classification, each node $v$ has a label $y_v \in \mathcal{Y} = \{1,2,...,Y \}$. Then, the loss function can be defined using the decoding function $DEC(\cdot)$ and Softmax:
\begin{align}
\ell = \frac{1}{N}\sum_{v \in \mathcal{V}} \text{Softmax}(DEC(\mathbf{h}_{v}),y_v)
\end{align}

In an unsupervised scenario, a cross-entropy loss can be used to encourage similarities between nearby nodes and difference between disparate nodes \cite{hamilton2017inductive}:
\vspace{-5pt}
\begin{align}
\ell = \log(\sigma(\mathbf{h}_{u}^{T} \mathbf{h}_{v})) + \sum_{n=1}^K \mathbb{E}_{v_n \sim P_n(v)} \log (\sigma(-\mathbf{h}_{v_n}^{T} \mathbf{h}_{v}))
\end{align}
where $u \in \mathcal{N}_v$ is a neighbor of $v$, $P_{n(v)}$ is the noise distribution for negative sampling, $K$ is the number of negative samples, and $\sigma$ denotes the sigmoid function.  

\medskip
\noindent \textbf{Problem Statement}. Suppose we have a graph $G=(\mathcal{V}, \mathcal{E}, \mathbf{X})$ and a well-trained $f_{GNN}$ on $G$. In the testing phase, $f_{GNN}$ is expected to perform good predictions on $G$ (\textit{e.g.}, transductive classification) and on other graphs with a similar distribution of $G$ (\textit{e.g.}, inductive classification). The goal of an {\em adversarial attack} is to maliciously transform $G$ to a perturbed graph $G'=(\mathcal{V}', \mathcal{E}', \mathbf{X}') = P(G)$, such that the performance of the original model $f_{GNN}$ on the perturbed graph $G'$ drops dramatically.

In this work, we study the problem of {\em adversarial defense} for GNN, that is to defend a GNN against the adversarial attacks. More formally, given a graph $G$, our goal is to build a robust GNN framework $f_{GNN}^{(robust)}$ during the training phase, such that in the testing phase, the robust model preserves its good performance on the attacked $G'$.

\vspace{-3pt}
\section{The \defmodel Framework}
\label{sec:method}

\begin{figure*}[!htb]
\centering
\vspace{-10pt}
\includegraphics[width =0.9\textwidth]{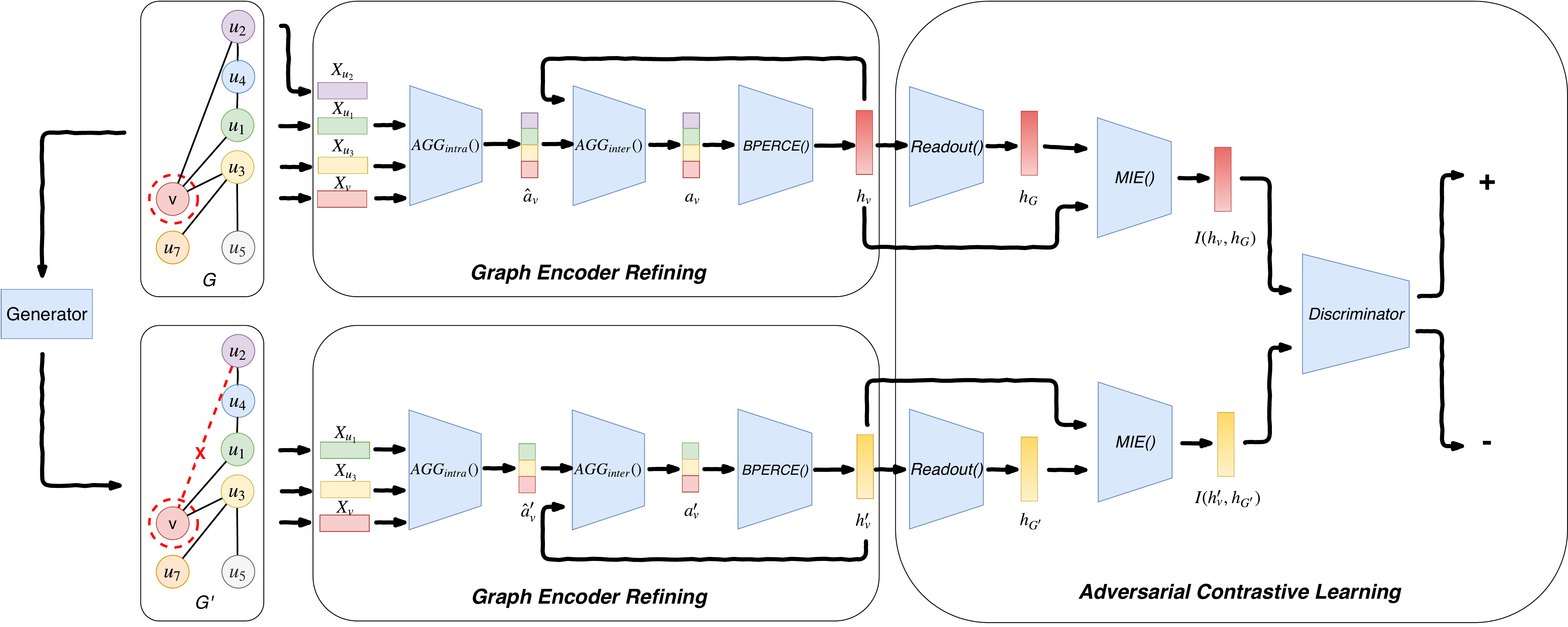}
\caption{An overview of \defmodel framework. \defmodel contains two modules: Graph Encoder Refining (\textbf{GER}) and Adversarial Contrastive Learning (\textbf{ACL}). The \textbf{GER} module polishes the encoder of GNN with mean intra-layer aggregation (\textit{i.e.}, \textit{AGG$_{intra}()$}), dense-connected inter-layer aggregation (\textit{i.e.}, \textit{AGG$_{inter}()$}) and bottleneck perceptron (\textit{i.e.}, \textit{BPERCE()}). The \textbf{ACL} module takes the benign graph samples (\textit{e.g.}, $G$) and the adversarial graph samples (\textit{e.g.}, $G'$) as the input to train the refined GNN in a conditional GAN manner.}
\label{fig:Architecture}
\vspace{-9pt}
\end{figure*}

To address the two key challenges introduced in Section~\ref{sec:intro}, we propose an adversarial defense framework \defmodel with two modules: Graph Encoder Refining (\textbf{GER}) and Adversarial Contrastive Learning (\textbf{ACL}) as illustrated in Fig.~\ref{fig:Architecture}. In \ger module, we aim to examine the vulnerabilities in every layer of a \gnn encoder, and propose corresponding strategies to address these vulnerabilities. And in \acl module, our goal is to train the \gnn to be more distinguishable between real benign samples and adversarial samples.

\subsection{GER: Graph Encoder Refining}
\nop{The robust graph encoder aims to embedded the graph in to a low-dimensional nonlinear space, which is robust to adversarial attacks. Two questions arise}
\nop{In this section, we address two key questions: (1) how to aggregate the contexts in a safe way; and (2) how to map the aggregated representation into a latent space which is robust to adversarial samples.}\nop{Can be cut if needed}

\subsubsection{Dual-Stage Aggregation}
As the first layer of GNN, the aggregation layer leverages the graph structure (\textit{i.e.}, the local context of each node) to improve the \gnn performance by aggregating neighbourhood. However,\nop{it is also an influential vulnerability,} due to graph properties such as homophily \cite{london2014collective},\nop{such that} the representation of a single node can easily be affected by its local context.\nop{\zach{What do you mean here}}\nop{Under this graph effects,} Consequently, attackers can compromise a node's prediction without directly changing its features and/or edges \cite{zugner2018adversarial}. \nop{Aggregation module improve the performance of the classification, but open a backdoor to the indirect adversarial attack.} \nop{It requires the aggregator to be robust to adversarial attack.}
\nop{As a GNN, the first component is the neighbourhood aggregator, which aims to generate the aggregated node representation by aggregating over local node neighbourhoods in the neighbourhood set $NE(v)$. This aggregation operation is similar to convolutional operation in a receptive field in computer vision. The choice of aggregator are very important and determines the robustness of GNNs.}

To improve the robustness of the aggregation layer, we propose a dual-stage aggregator. As shown in the middle of Fig. \ref{fig:Architecture}, in the first stage, an intra-layer aggregation integrates the neighbourhood in a \textit{mean pooling} manner. In the second stage, an inter-layer aggregation combines the node representations from multiple layers in a \textit{dense-connection} manner.

The intra-layer aggregation aims to compute the new representation of a node by aggregating its local neighbourhood spatially.\nop{In this way, the graph structure are embedded during the learning procedure.}
Typical neighbourhood aggregation operation includes sum \cite{xu2018powerful}, max \cite{hamilton2017inductive}, and mean \cite{kipf2016semi}.\nop{They are permutation-invariant and can take the neighbourhood without any order information.}
The sum aggregator sums up the features within the neighbourhood $\mathcal{N}_{v}$, which captures the full neighbourhood.
The max aggregator generates the aggregated representation by element-wise max-pooling. It 
captures neither the exact structure nor the distribution of the neighbourhood $\mathcal{N}_{v}$.
The mean aggregator averages individual element features out. Different from the sum and max aggregators, it can capture the distribution of the features in the neighbourhood $\mathcal{N}_v$. 

The adversarial attacks are usually stealthy and can only perform small-scale modifications to the graph data.
An ideal aggregator should be robust to such subtle perturbations. Among the three aggregators, the max aggregator is very sensitive to the distinct modification  (\textit{e.g.}, by adding a new neighbor of a different class with a big value), even though it is very small. 
For the sum aggregator, it can also be affected by this kind of modification.
In contrast, the mean aggregator is less sensitive to the small-scale modification and thus is more robust to adversarial attacks.

\nop{\shen{maybe add motivated example figures as ICLR paper}}

Therefore, in our \textbf{GER} module, the intra-layer aggregation $AGG_{intra}()$ aggregates the neighborhood in a mean pooling manner. Formally, it is defined as follows:

\begin{align}
\hat{\mathbf{a}}_v^{(k)} & = AGG_{intra}( \{\mathbf{h}_u^{(k-1)}:u \in \mathcal{N}_v \}))\\
         & = \frac{1}{|\mathcal{N}_v|} \sum_{u \in \mathcal{N}_v} \mathbf{h}_u^{(k-1)}
\end{align}
where $k$ is the index of layer/iteration, $|\mathcal{N}_v|$ is the cardinality of the neighborhood of node $v$, and $\hat{\mathbf{a}}_v^{(k)}$ is the result of the intra-layer aggregation.

In the second stage, the layer-wise aggregation is employed to connect current layer of the network to its previous layers, such that more structural information can be covered. Recently, GNNs, such as GCN and GraphSAGE, leverage the \textbf{skip-connection} to aggregate the representation from the predecessor layer and drop the rest of intermediate representations during the information propagation within $k$-hop neighbors. However, stacking multiple such layers could also propagate the noisy or adversarial information from an exponentially increasing number of expanded neighborhood member \cite{kipf2016semi}.
\nop{\shen{However, stacking multiple such layers 
would be sensitive to adversarial attack
and the effect of adversarial attack will be accumulated until the last layer, which would eventually hurt the robustness of the prediction.}
\shen{This is mainly because more
layers could also propagate the noisy information from
an exponentially increasing number of expanded neighborhood member \cite{kipf2016semi}. Similarly, the effect of adversarial attack will be accumulated until the last layer.}}
\nop{These drawbacks make GNN 
sensitive to the adversarial attacks and thus easy to be compromised.}

To address this problem, we propose a \textbf{dense-connected} inter-layer aggregation, as inspired by the DenseNet~\cite{huang2017densely}. Our method keeps all intermediate representations and aggregates them together to compute the recent-layer representation. In this way, the recent layer is connected with all previous hidden layers, allowing the subsequent layer to selectively but adaptively aggregate structure information from different hops. Consequently, the robustness can be improved for deep GNN.

Formally, the \textbf{dense-connected} inter-layer aggregation $AGG_{inter}()$ is constructed as follows:
\begin{align}
\mathbf{a}_v^{(k)} & = AGG_{inter}(\hat{\mathbf{a}}_v^{k}, \mathbf{h}_v^{(k-1)}, \mathbf{h}_v^{(k-2)}... \mathbf{h}_v^{(1)})\\
                & = [\hat{\mathbf{a}}_v^{k}; \mathbf{h}_v^{(k-1)}; \mathbf{h}_v^{(k-2)}; ...; \mathbf{h}_v^{(1)}]
\end{align}
where $[\cdot; \cdot]$ represents the feature concatenation operation. 

\subsubsection{Bottleneck Perceptron}
In this section, we propose the bottleneck perceptron for mapping the aggregated representation to node embedding in a non-linearly low-dimensional space. Particularly, it consists of a bottleneck mapping followed by a nonlinear mapping. 

Recent researches have 
empirically studied the one-to-one relationship between the adversarial vulnerability and the input dimensionality, and theoretically proved that the adversarial vulnerability of neural networks deteriorates as the input dimensionality increases \cite{simon2018adversarial}. 
On the other hand, 
it is common that real data are not truly high-dimensional \cite{levina2005maximum}. The data in a high-dimensional space can be well embedded into a lower dimensional space for both better effectiveness and efficiency. Therefore, 
it is important to perform dimensionality reduction on the input to improve the robustness of a \gnn.

\nop{In this module,}To this purpose, we design a bottleneck mapping, such that the output dimensionality is much lower than that of the input. Meanwhile, we add a nonlinear gating function to capture the nonlinearity of the data.\nop{When we construct the deep network structure, we can benefit universal approximation theorem to get GNNs we need.} In particular, we use rectified linear unit (ReLU) as the 
activation function. Formally, the bottleneck perceptron is defined as follows:
\begin{align}
\mathbf{h}_{v}^{(k)} = BPERCE (\mathbf{a}_{v}^{(k)})
          = RELU ((\mathbf{W}^{(k)}_{p}) \mathbf{a}_v^{(k)})
\end{align}
where $\mathbf{W}^{(k)}_p$ is the trainable mapping matrix. The output of this module is the low-dimensional node embedding.

\vspace{-2pt}
\subsection{ACL: Adversarial Contrastive Learning}
\nop{Even with the help of the proposed dual-stage aggregation and bottleneck perceptron,}After addressing the vulnerabilities of the graph encoder, the \gnn may still be vulnerable to the adversarial attacks due to the scarcity of training data~\cite{pmlr-v70-pinto17a}. To 
handle this issue, in this section, we introduce the adversarial contrastive learning to regularize the training of the GNN.

Contrastive learning has been widely found useful for representation learning of graph data \cite{hamilton2017representation}. Its objective is written by:
\begin{align}
\label{eq:contrast}
L &=\mathbb{E}_{(x^+,y^+,y^-)}[\ell_{\mathbf{w}}(x^+,y^+,y^-)]
\end{align}
where $\ell_{\mathbf{w}}(x^+,y^+,y^-)$ scores a positive tuple $(x^+,y^+)$ against a negative one $(x^+,y^-)$, and $\mathbb{E}_{(x^+,y^+,y^-)}$ represents the expectation 
w.r.t. the joint distribution over the positive and negative samples. Since negative sampling is independent of the positive label, Eq. \ref{eq:contrast} can be rewritten by:
\begin{align}
L &=\mathbb{E}_{p(x^+)}\mathbb{E}_{p(y^+|x^+)p(y^-|x^+)}[\ell_{\mathbf{w}}(x^+,y^+,y^-)]\\
&=\mathbb{E}_{p(x^+)}[\mathbb{E}_{p^+(y|x)}D_{\mathbf{w}}(x,y) - \mathbb{E}_{p^-(y|x)}D_{\mathbf{w}}(x,y)]
\end{align}
where $D_{\mathbf{w}}(x,y)$ measures the correlation between $x$ and $y$.

Traditional graph embedding methods adopt a noise contrastive estimation approach and approximate $p(y^-|x^+)$ with a noise distribution probability $p_n(y)$, such that 
\begin{align}
L &=\mathbb{E}_{p(x^+)}[ \mathbb{E}_{p(y^+|x^+)p_n}[\ell_{\mathbf{w}}(x^+,y^+,y^-)]]\\
&=\mathbb{E}_{p(x^+)}[\mathbb{E}_{p^+(y|x)} D_{\mathbf{w}}(x,y) - \mathbb{E}_{p_n} D_{\mathbf{w}}(x,y)]
\end{align}

However, using $p_n$ sacrifices the performance of learning as the negative samples are produced based on a noise distribution. What is worse, using $p_n$ also harms the robustness of the learning as the negative samples are produced without considering the adversarial samples. 

To address this problem, we model the negative sampler by a generator under a conditional distribution $g_{\theta}(y|x)$. Formally, it is defined as follows:
\begin{align}
L &=\mathbb{E}_{p(x^+)}[\mathbb{E}_{p^+(y|x)} D_{\mathbf{w}}(x,y) -\mathbb{E}_{g_{\theta}(y|x)} D_{\mathbf{w}}(x,y)].
\end{align}
We optimize the above objective function in a minimax adversarial learning manner as follows:\nop{to enhance the robustness of the model. The objective is shown a follows:} 
\vspace{-3pt}
\begin{dmath}
\min_{G_{\theta}} \max_{D_{\mathbf{w}}} 
 \{ \mathbb{E}_{p^+(y|x)} [ \log D_{\mathbf{w}}(x,y)]
 + \mathbb{E}_{g_{\theta}(x,y)} [ \log(1-D_{\mathbf{w}}(G_{\theta}(x,y)))] \}.
\end{dmath}

This formulation is closely related to\nop{\zach{"closely related to" or "close to" }} GAN by regarding $D_{\mathbf{w}}$ as the discriminator and $G_{\theta}$ as the generator. The generative model $G_{\theta}$ captures the data distribution, and the discriminative model $D_{\mathbf{w}}$ estimates the probability that a sample comes from the training data rather than $G_{\theta}$.
$G_{\theta}$ and $D_{\mathbf{w}}$ are 
trained jointly: we adjust the parameters of $G_{\theta}$ to minimize $\log(1-D_{\mathbf{w}}(G_{\theta}(x,y)))$
and adjust the parameters of $D_{\mathbf{w}}$ to minimize $\log D_{\mathbf{w}}(x,y)$, as if they follow the two-player minimax game.

However, the vanilla GAN does not take any constraint into account for the generator, and the generator is not tailored to the graph data. 
This may cause suboptimal quality of the generated adversarial samples since the adversarial latent space can not be effectively explored.
This limitation may reduce the performance of the feature learning. To overcome it, we adopt a conditional GAN configuration, which extends a conditional model where the generator and discriminator are both conditioned on some extra information. The extra information could be any kind of auxiliary information, such as class labels or data from other modalities.

The condition 
can be realized by feeding extra information into both the discriminator and generator as an additional input layer. As shown in Fig.~\ref{fig:Architecture}, we consider the high-level graph representation (\textit{e.g.}, graph embedding $\mathbf{h}_G$) as 
a global constraint to regularize the node representation learning. 
The motivation is based on the observation \cite{dai2018adversarial} that global level representation such as graph embedding is less sensitive to the adversarial perturbation than the local-level representation such as node embedding. Thus, the global graph representation can be used to guide the robust learning of the local node representations. Accordingly, the objective of our adversarial contrastive learning becomes:
\vspace{-3pt}
\begin{dmath}
\min_{G} \max_{D} 
 \{ \mathbb{E}_{p^+} [ \log D_{\mathbf{w}}(\mathbf{h}_v|\mathbf{h}_G)] \\
 + \mathbb{E}_{g_{\theta}} [ \log(1-D_{\mathbf{w}}(G_{\theta}(\mathbf{h}'_v|\mathbf{h}_G))))] \}
\end{dmath}
where $\mathbf{h}'_v$ is the node embedding 
from the generated adversarial samples $G'=(V',E',X')=P(G)$. Here, the perturbation function $P(\cdot)$ can be\nop{a trained attack model as in \cite{zugner2018adversarial}.} defined similarly to \cite{zugner2018adversarial}.

\begin{figure*}[t]
\vspace{-15pt}
\centering
\includegraphics[width =0.81\textwidth]{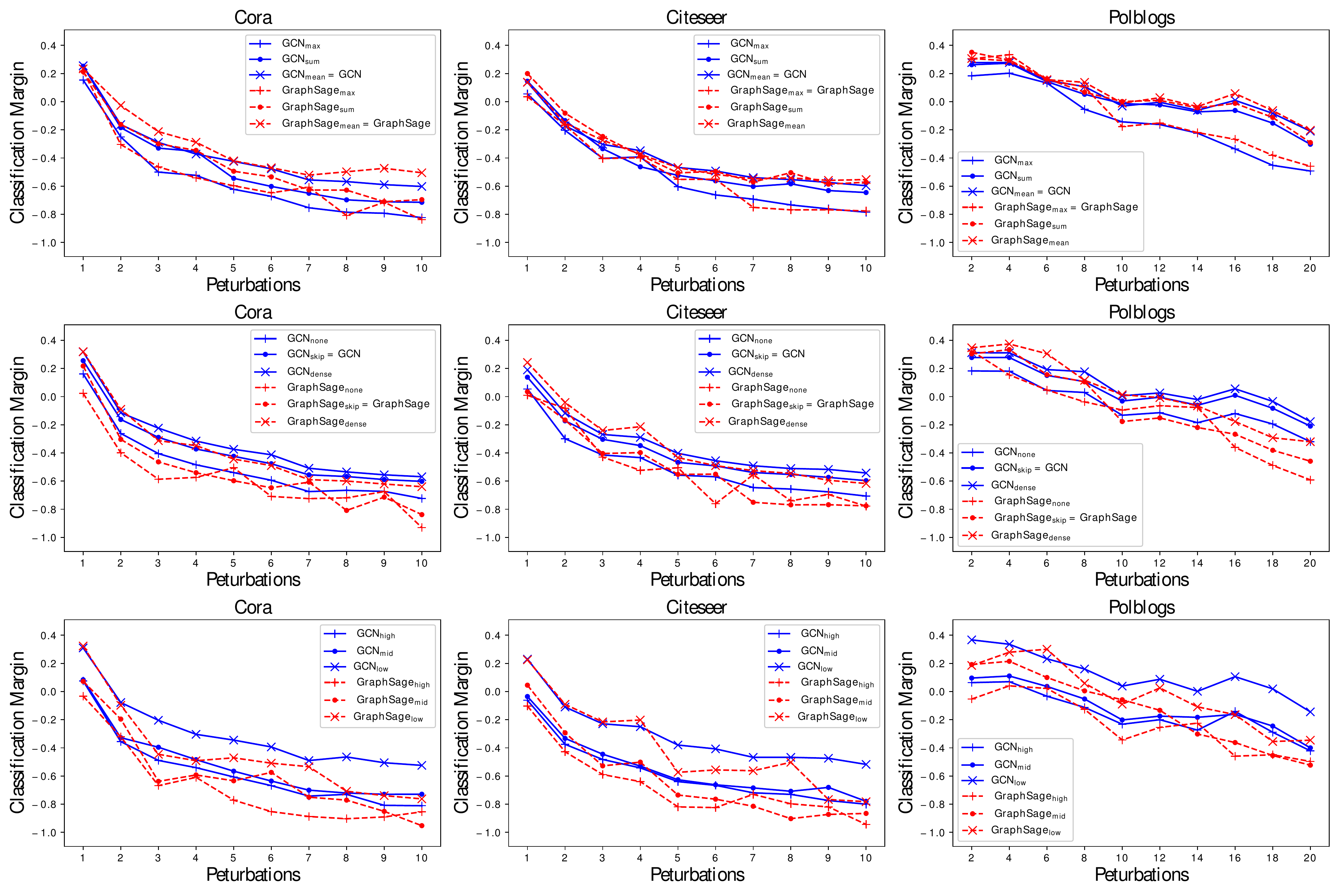}
\vspace{-4pt}
\caption{Evaluation of the \textbf{GER}.}
\label{fig:exp_1}
\vspace{-10pt}
\end{figure*}

To obtain the graph embedding, we employ a $READOUT$ function. Given the node representations $\{h_v\}_{v \in \mathcal{V}}$ of graph $G$, the $READOUT$ function performs a mean pooling followed by a nonlinear mapping as follows:
\begin{align}
\mathbf{h}_G &= READOUT ({\mathbf{h}_v^{(k)} | v \in \mathcal{V}}) \\
    &= \sigma (\mathbf{W}_r (\frac{1}{N} \sum_{i=1}^N \mathbf{h}_i^{K})))
\end{align}
where $\sigma$ represents the sigmoid 
function, $\mathbf{W}_r$ represents a trainable linear mapping matrix, and $K$ denotes the total number of layers/iterations. 
This module follows the idea of using the pooling function to compute the high-level global representation in CNN. 
The reason of using a mean pooling is similar to that of neighborhood aggregation, \textit{i.e.}, the representation generated by the mean operation is more stable and less sensitive to the small-scale adversarial attacks.

For the discriminator, we employ a mutual information estimator to jointly model the local and global graph representations via a bilinear scoring function: 
\begin{align}
\vspace{-5pt}
D_{\mathbf{w}}(\mathbf{h}_v|\mathbf{h}_G) 
&=I(\mathbf{h}_v,\mathbf{h}_G)=\sigma(\mathbf{h}_v \mathbf{W}_{D} \mathbf{h}_G)
\vspace{-5pt}
\end{align}
where $I(\mathbf{h}_v,\mathbf{h}_G)$ represents the mutual information between the node embedding and graph embedding. $\mathbf{W}_{D}$ is a trainable scoring matrix and $\sigma$ denotes the sigmoid 
function. 

\vspace{-3pt}

\section{Experiments}
\label{sec:exp}
In this section, we evaluate our proposed \defmodel framework on defending adversarial attacks. 
\vspace{-2pt}
\subsection{Datasets}

\begin{table}[!htp]
\centering
\small
\caption{The statistics of the datasets.}
\label{tab:dataset}
\vspace{-5pt}
\begin{tabular}{lllll}
\hline
Dataset  & \#Node  & \#Edge  & \#Feature & \#Class                      \\ \hline
Cora     & 2,810   & 7,981   & 1,433 (Categorical)          & 7                                    \\
Citeseer & 2,110   & 3,757   & 3,703 (Categorical)         & 3                                    \\
PolBlogs & 1,222   & 16,714  & Identity feature & 2  
\\ \hline
\end{tabular}
\vspace{-8pt}
\end{table}

\nop{In the experiments,}
We use three benchmark datasets including two academic networks (Cora and Citeseer)\nop{with categorical node features} and a social network (PolBlogs) for node classification tasks.\nop{without features}\nop{We construct the one-hot identity node feature for PolBlogs.} Table~\ref{tab:dataset} shows the statistics of these datasets. We select the largest connected components in each dataset for our experiments, and we split all the datasets randomly into a labeled set (80\%) and an unlabeled set (20\%). We then further divide the labeled set into a training set (50\%) and a validate set (50\%). 
\nop{The labels of the validation set are removed during the training and the validation set are used as stop criterion. During the testing stage, we conduct the evaluation on the unlabeled set.}

\subsection{Experiment Setup}
\nop{In this section, we introduce the adversarial attack model, baseline method and experiment configuration used in the experiments. }
\subsubsection{Attack Models}
\nop{In this part,The adversarial attack model used in the experiments are introduced. Specifically,We consider a more real-world setting -- black-box attacks in our experiments. Three types of adversarial attacks are conducted: \nop{including a random perturbation attack (\textbfRAND), a gradient sign method (\textbfFGSM) \cite{goodfellow2015explaining} and a state-of-the-art method NETTACK \cite{zugner2018adversarial}.}}
We conduct three types of popular adversarial attacks:\nop{in a black-box setting, which is more realistic}
\begin{itemize}
\vspace{-2pt}
\item \textbf{RAND} (a random perturbation attack):\nop{is a random perturbation attack.} Given the target node, it randomly adds edges to the node that belongs to a different class and/or deletes an edge that connected to the node within the same class.
\vspace{-2pt}
\item \textbf{FGSM} (a gradient based attack)~\cite{goodfellow2015explaining}):\nop{is a gradient based attack method, which} It generates adversarial examples based on the sign of gradient.\nop{Since we consider the black box attack setting, FGSM is performed on a surrogate model -- a linear GNNs as used in \cite{zugner2018adversarial}.} 
\vspace{-2pt}
\item \textbf{NETTACK} (a state-of-the-art optimization based attack~\cite{zugner2018adversarial}):\nop{is an optimization based attack model.} It generates adversarial perturbations by searching the perturbation space.\nop{, which can lower the confidence of prediction. This attack model is also applied to a surrogate model. We only consider the structure attack setting, which perturbation is only performed on the graph structure.}
\end{itemize}
The attack procedure is similar to \cite{zugner2018adversarial}.\nop{but the number of attacked node is extended,} We select $200$ nodes as the targets: $50$ correctly classified nodes with high confidence; $50$ correctly classified nodes but with low confidence; and $100$ random nodes. 

\vspace{-1pt}
\subsubsection{\textbf{Comparing Methods}}
\nop{To show the venerability of GNN and the effectiveness of the proposed method against the adversarial attacks,We compare our \defmodel with a number of baselines.}
We evaluate our framework \defmodel on two most popular \gnn variants: GCN \cite{kipf2016semi} and GraphSAGE \cite{hamilton2017inductive}.\nop{We study the venerability of GNN and test the effectiveness of proposed graph encoder refiner on two most popular Graph convolutional networks: Graph Convolutional Network (GCN) \cite{kipf2016semi} and GraphSAGE \cite{hamilton2017inductive}. In this group of experiments, we consider GNN variants in terms of intra-layer aggregator} To evaluate the \ger module, we compare different versions of GCN and GraphSAGE in terms of intra-layer aggregators (XXX$_{mean}$, XXX$_{sum}$, XXX$_{max}$), inter-layer aggregators (XXX$_{none}$ (without inter-layer aggregator), XXX$_{skip}$ (skip based), XXX$_{dense}$ (dense based)), and output dimensions (XXX$_{low}$, XXX$_{mid}$, XXX$_{high}$) as discussed in Section \ref{sec:method}. To evaluate the \acl module, we use the refined GNN after \textbf{GER} (RGCN or RGraphSage) as graph encoder, and consider their variants in terms of adversarial learning methods:
XXX$_{NCL}$ (noise contrastive learning) and XXX$_{ACL}$ (adversarial contrastive learning).
\vspace{-1pt}
\subsubsection{\textbf{Configuration}}
We repeat all the experiments over five different splits of labeled/unlabeled nodes to compute the average results. For GNNs used in the experiments, we set the default number of layers as $3$, and the number of perceptron layers as $2$. We also adopt the Adam optimizer with an initial learning rate of $0.01$, and decay the learning rate by $0.5$ for every $50$ epochs.

\subsection{Evaluation of \textbf{GER}}
In this experiment, we show the vulnerabilities of the GNN in terms of its aggregation and perceptron components. We also demonstrate the effectiveness of \ger in addressing these vulnerabilities.

\nop{Specifically, we take the most popular GNN model GCN \cite{kipf2016semi} as baseline, which adopts \underline{mean intra-layer aggregator} and \underline{simple inter-layer aggregator} (aggregation of both target node and context nodes). To test and analysis the venerability of aggregation and perceptron components, We consider three types variants, including type of intra-layer aggregator, type of inter-layer aggregator and dimension of the output for perceptron layer. To test the robustness of intra-layer aggregator, we construct $GCN_{sum}$ and $GCN_{max}$, which performs sum aggregation and max aggregation. To test the robustness of inter-layer aggregator, we construct $GCN_{none}$, $GCN_{skip}$, which performs no intra-layer aggregation and skip inrea-layer aggregation. To test the robustness of the last layer dimension of the perceptron module, we construct $GCN_{low}$, $GCN_{mid}$ and $GCN_{high}$. The proposed REG adopts a \underline{mean intra-layer aggregator} and \underline{skip inter-layer aggregator}. Similarly, we also consider its three type variants.} 
We test the robustness of different methods under the most powerful attack NETTACK with various adversarial perturbations. We use the classification margin score $s$ as the metric for the classifier performance: \nop{which is defined as follows:}
\vspace{-5pt}
\begin{align}
s=Z^*_{v,c^{gt}_v}-max_{c\neq c^{gt}_v} Z*_{v,c}
\end{align}
where $Z^*_{v,c^{gt}_v}$ is the predicted probability for the ground-truth label, while  $max_{c\neq c^{gt}_v} Z*_{v,c}$ is the incorrect prediction probability with the largest confidence. The larger $s$ is, the higher confidence the classifier has in making a correct prediction. \nop{better, which indicates that the classifier can make the correct prediction with high confidence.}\nop{When this score is under zero, the classifier makes wrong prediction.} 

\nop{The experiment results are shown in Fig.~\ref{fig:exp_1}, with each row corresponding to one graph encoder component and each column corresponds to one dataset.} From the experimental results shown in Fig.~\ref{fig:exp_1}, we can see that the GCN and GraphSAGE with our proposed \ger perform better than the original version and the ones using other aggregation or perceptron methods.
\begin{table}[!htp]
\vspace{-5pt}
\caption{Performance comparison under various attacks.}
\label{tab:exp2}
\tiny
\centering
\begin{tabular}{clcccc}
\hline
\multirow{2}{*}{Data}     & \multirow{2}{*}{Defense} & \multirow{2}{*}{Clean} & \multicolumn{3}{c}{Attack models} \\
                          &                          &                        & RAND    & FGSM    & NETATTACK    \\ \hline
\multirow{8}{*}{Cora}   &$\textsc{GCN}$        &0.90     &0.60      &0.03      &0.01 \\
                        &$\textsc{RGCN}$       &0.88     &0.73      &0.18      &0.15 \\
                        &$\textsc{RGCN}_{NCL}$ &0.85     &0.80      &0.42      &0.38 \\
                        &$\textsc{RGCN}_{ACL}$ &\bf{0.92}&\bf{0.85} &\bf{0.65} &\bf{0.60} \\ \cline{2-6}  
                        &GraphSage  &0.85     &0.70      &0.18      &0.16 \\
                        &RGraphSage &0.88     &0.78      &0.25      &0.22 \\
                  &RGraphSage$_{RCL}$ &0.84     &0.82      &0.48      &0.42 \\ 
                  &RGraphSage$_{ACL}$ &\bf{0.88}&\bf{0.84} &\bf{0.67} &\bf{0.60}\\
                        \hline
\multirow{8}{*}{Citeseer}&$\textsc{GCN}$       &0.88     &0.60      &0.07      &0.02 \\
                        &$\textsc{RGCN}$       &\bf{0.90}&0.72      &0.20      &0.16 \\
                        &$\textsc{RGCN}_{NCL}$ &0.86     &0.79      &0.52      &0.45 \\
                        &$\textsc{RGCN}_{ACL}$ &\bf{0.90}&\bf{0.85} &\bf{0.70} &\bf{0.65} \\ \cline{2-6}
                        &GraphSage  &0.83     &0.70      &0.10      &0.04 \\
                        &RGraphSage &0.86&0.80      &0.22      &0.18 \\
                  &RGraphSage$_{RCL}$ &0.82     &0.82      &0.50      &0.48 \\ 
                  &RGraphSage$_{ACL}$ &\bf{0.88}     &\bf{0.84} &\bf{0.68} &\bf{0.65} \\
                        \hline
\multirow{8}{*}{Polblogs}&$\textsc{GCN}$       &0.93     &0.36      &0.41      &0.06\\
                        &$\textsc{RGCN}$       &\bf{0.95}&0.40      &0.50      &0.18 \\
                        &$\textsc{RGCN}_{NCL}$ &0.87     &0.65      &0.58      &0.52 \\
                        &$\textsc{RGCN}_{ACL}$ &\bf{0.95}&\bf{0.80} &\bf{0.74} &\bf{0.65} \\ \cline{2-6}
                        &GraphSage  &0.86     &0.43      &0.40      &0.14 \\
                        &RGraphSage &\bf{0.90}&0.60      &0.48      &0.20 \\
                  &RGraphSage$_{RCL}$ &0.82     &0.68      &0.59      &0.55 \\ 
                  &RGraphSage$_{ACL}$ &0.89     &\bf{0.78} &\bf{0.72} &\bf{0.68} \\
                        \hline
\end{tabular}

\vspace{-5pt}
\end{table}

We can also observe that: 
(1) The mean intra-aggregator outperforms the sum and max ones\nop{, which produces a robust and stable prediction};
(2) The dense inter-aggregator outperforms the case without the inter-aggregator and the skip inter-aggregator;\nop{. Furthermore, it helps to generate a stable prediction, even with a random sampling aggegator}
(3) The lower dimension of the perceptron layer beats both the high and medium ones;
(4) Comparing GCN with GraphSAGE, the performance of the latter one is less stable but sometimes performs better, due to a max aggregator combined with the random neighbourhood sampling; and (5) It is also worth noticing that PolBlogs is difficult to attack, since the average degree of the node is high.


\subsection{Evaluation of \textbf{ACL} Module and \defmodel Framework}
In this experiment, we show the effectiveness of the proposed \textbf{ACL} module and overall \defmodel framework under various attacks.\nop{against the various type of adversarial attacks, including RAND, FGSM and NETTACK. }
\nop{By the observation in the previous experiment, the attack's effect will be small if the number of the perturbation is smaller than the average node degree.} To guarantee the powerfulness of the adversarial attacks, we set the number of the perturbations as: $n_p=d_{v}+2$ with $d_{v}$ to be the degree of the target node. We report the fraction of the target nodes that get correctly classified as accuracy. The results in Table~\ref{tab:exp2} show that the proposed \acl method outperforms all the baselines including the traditional training and noise contrastive learning in terms of datasets, GNN variants, and attack models. We also observe that the original GCN and GraphSAGE are both very venerable to the attacks,  but after applying our \defmodel framework, both of them become way more robust. For example, GCN can achieve an accuracy of $0.90$ without any attack on Cora data, but the performance significantly drops to $0.03$ under the \textbf{FGSM} attack, while RGCN$_{ACL}$ (\textit{i.e.}, GCN with \defmodele) can still achieve an accuracy of $0.65$.    

\nop{More specifically: 
(1) The \acl outperform the traditional label supervised end-to-end training baselines;
(2) The \acl beats the noise contrastive learning method;
(3) Adversarial contrastive learning methods outperform all the baseline method even in a clean case, which can help to improve both the robustness and performance of the prediction. }


\vspace{-3pt}
\section{Related Work}
\nop{In line with the focus of this work, we briefly review the recent work on graph neural network, adversarial attack and defense on traditional machine learning, and adversarial attack and defense on graph neural networks.}

\subsection{Graph Neural Networks}
In recent years, representation learning on graph data via Graph Neural Network (GNN) \cite{defferrard2016convolutional,hamilton2017inductive,kipf2016semi,velickovic2017graph,li2015gated,gilmer2017neural,xu2018powerful} has attracted increasing interests.\nop{With the help of the powerful representation learning ability, \textbf{GNN}s have achieved the state-of-the-art results on different graph analysis tasks (including node classification and graph classification) and even on some other fields, such as computer vision, computer graphics, and nature language processing.
The basic idea of the GNN is to leverage the graph structure during the learning procedure.} Many GNN variants have been proposed to leverage the graph structure to capture different proprieties. \cite{defferrard2016convolutional,hamilton2017inductive,kipf2016semi,li2015gated} followed a neighbourhood aggregation scheme. \cite{gilmer2017neural} leveraged the structure information via message passing operation. Their success is built on a supervised end-to-end data-driven learning framework, which is vulnerable to adversarial attacks.

\subsection{Attack and Defense on \textbf{GNNs}}
Recently, a few attempts have been made to study adversarial attacks on \textbf{GNNs}. \cite{dai2018adversarial} proposed a non-target evasion attack on node classification and graph classification. A reinforcement learning method was used to learn the attack policy that applies small-scale modification to the graph structure. \cite{zugner2018adversarial} introduced a poisoning attack on node classification. This work adopted a substitute model attack and formulated the attack problem as a bi-level optimization. Different from \cite{dai2018adversarial}, it attacked both the graph structure and node attributes. Furthermore, it considered both direct attacks and influence attacks. However, there is still limited understanding on why GNNs are vulnerable to adversarial attacks and how to defense against the adversarial attacks.\nop{of how robust are graph neural network against the adversarial attack and there is no work about the adversarial defense against for GNNs.} This motivates our work. 

\vspace{-3pt}
\section{Conclusion}
In this paper, we presented \defmodele, an adversarial defense framework for improving the robustness of Graph Neural Networks (GNNs) under adversarial attacks.
To address the vulnerabilities in the aggregation layer and perceptron layer of GNN, we proposed the dual-stage aggregation and bottleneck perceptron methods. By leveraging the high-level graph representation, we proposed the adversarial contrastive learning technique to train the \gnn in a \textit{conditional GAN} manner. We evaluated the proposed framework using extensive experiments on three public graph datasets. The experimental results convince us of the effectiveness of our framework on defending the popular \gnn variants against various types of adversarial attacks.

\end{document}